\documentclass{article}
\usepackage[preprint]{spconf}
\usepackage{amsmath,graphicx,hyperref}

\title{Multi-Task Partially Supervised Learning for Super-Resolution and Semantic Segmentation on Earth Observation data} 

\name{
Hoàng-Ân Lê$^{1,2,*}$\thanks{$^*$Work done as part of the CNES post-doctoral fellowship program},
Minh-Tan Pham${}^{1}$,
Solange Lemai-Chenevier${}^{2}$, Daniel Greslou${}^{2}$ 
}
\address{
${}^1$Université Bretagne Sud, IRISA, UMR 6074, Vannes, France \\
${}^2$Centre National d’Etudes Spatiales (CNES), Toulouse, France
}

\usepackage{lipsum}
\usepackage{framed,multirow}
\usepackage{booktabs}
\usepackage{url}
\usepackage{xcolor}

\usepackage{comment}

\usepackage{algorithm}
\usepackage[noend]{algorithmic}
\usepackage{eqparbox}

\usepackage{graphicx}
\usepackage{amsmath,amssymb} %
\usepackage{bm}
\usepackage{booktabs}
\usepackage{rotating,multirow}
\usepackage{wrapfig}

\usepackage{tikz}
\usepackage{transparent} %
\usepackage{calc} %
\usepackage[export]{adjustbox}
\usepackage{bbm} %

\usepackage{soul} %

\usepackage{svg}

\input{_definitions}

\copyrightnotice{\parbox{\textwidth}{\footnotesize\centering\copyright\ 2026 IEEE. Personal use of this material is permitted. Permission from IEEE must be obtained for all other use, in any current or future media, including reprinting/republishing this material for advertising or promotional purposes, creating new collective works, for resale or redistribution to servers or lists, or reuse of any copyrighted component of this work in other works. DOI: 10.1109/ICIP61757.2026.11630424”
}}
\begin{document}
\maketitle

\begin{abstract}
Super-resolution and semantic segmentation are known to benefit one another, especially in the Earth observation context. However, learning both tasks in a joint model often requires both task annotations, which is impractical and expensive. In this paper, we study the multi-task partially supervised learning paradigm for both tasks, where each example is assumed to have only a single-task annotation. To that end, we examine two multi-task architectural variations, the sequential and shared variants, then propose a hybrid variant and a re-projection loss to benefit from the shared representation and enforce image quality of super-resolution when training with semantic segmentation. Experiments show favorable results compared to the SOTAs sequential variant. Source codes will be published at \url{https://github.com/lhoangan/munera}.

\end{abstract}

\begin{keywords}
super-resolution, semantic segmentation, multi-task, partially supervised learning
\end{keywords}
\section{Introduction}
\label{sec:intro}

Earth observation, which studies and monitors the characteristics and activities on the Earth's surface, has been attracting the attention of the computer vision community because of its importance and potential. %
One prominent challenge of Earth observation data is the high variation of data appearances due to geographical differences,~\eg rural compared to urban areas, and due to different capturing sensors and acquisition altitudes,~\eg aerial and satellite platforms. Scaling variation also affects scenes' detail level, which is generally crucial for semantic-based applications~\cite{Toker2024,Kim2024SR4IR}.

Super-resolution, which upsamples images and recovers high-resolution details, could serve as a remedy.
However, it is an ill-posed problem, as there exists, in principle, an infinite number of high-resolution images that, through different degrading processes, result in the same low-resolution input image~\cite{BellKligler19}.
Categorical information has proven effective in constraining super-resolution~\cite{Aakerberg2022}, allowing realistic textural reconstruction~\cite{Wang2018texture,Rad2020} and improving photo-realistic qualities.

As it is known that the multi-task paradigm learns domain-specific information from related tasks to improve one another, jointly learning super-resolution and semantic segmentation in a single model should be reciprocally beneficial for Earth observation applications.
The dual-task model by Wang~\etal~\cite{Wang2020DSRL} with a shared encoder and two task-heads enhancing representation detail levels and predicting semantic segmentation at larger resolutions has been applied to remote sensing by various authors~\cite{Abadal2021,Zhang2022colnet} while being adapted to different network architectures~\cite{Salgueiro2022}. On the other hand, \cite{Rad2020} and \cite{Aakerberg2022} jointly learn semantic segmentation and super-resolution for the super-resolution's benefit.
Although various approaches in the literature have leveraged one task to improve the performance of the other ~\cite{Wang2018texture,Rad2020,Wang2020DSRL,Abadal2021}, focusing on both tasks simultaneously~\cite{Zhang2022colnet,Salgueiro2022} remains limited, due to the impractical requirements of annotations for all tasks per training example that multiply the effort in creating a dataset and maintaining task synchronization.

Multi-task partially supervised learning~\cite{Li2022MTPSL}, in which each training example is required to be annotated for only a single task, is proposed to relax the strict requirement of multi-task learning. Without requiring all-task annotations for each training example, multiple datasets already annotated for different tasks can be combined in a single training pipeline, increasing the number of training examples and benefiting from complementary information. Yet, it is not without hindrances. The lack of joint annotations hinders joint optimization during training and requires specific attention regarding the tasks and input data~\cite{Le2023BMVC}.
In this paper, we jointly learn super-resolution and semantic segmentation under the same network with a shared encoder using partially annotated datasets for Earth observation. To that end, we propose a re-projection loss to constrain the super-resolution subnet when training with segmentation-annotated data.

Despite several works on discriminative tasks such as depth, segmentation, and object detection, the combination of a generative (super-resolution) and a discriminative (semantic segmentation) task is still limited, especially for Earth observation. Similar to our setting are the semantic-guided super-resolution~\cite{Aakerberg2022} and super-resolution for image recognition (SR4IR)~\cite{Kim2024SR4IR}, which places a semantic segmentation network after the super-resolution network in a sequential order. Although such a design orients super-resolution toward a downstream task, it places semantic segmentation under the entire influence of super-resolution performance and reduces shared representation. To make the best of both worlds, we propose a hybrid variant that allows shared features as well as adapting super-resolution for a downstream task. The experiments performed on the ISPRS~\cite{isprs2012} and proprietary Pleiades-based datasets show favorable results compared to the existing state-of-the-art.

The contributions of the paper are as follows: We (1) study the multi-task partial supervised learning for super-resolution and semantic segmentation on Earth observation data, and (2) propose a hybrid variant and a re-projection loss to enforce super-resolution image quality.

    \begin{figure*}[t]
        \centering
        \def\svgwidth{\textwidth}
        \scriptsize
        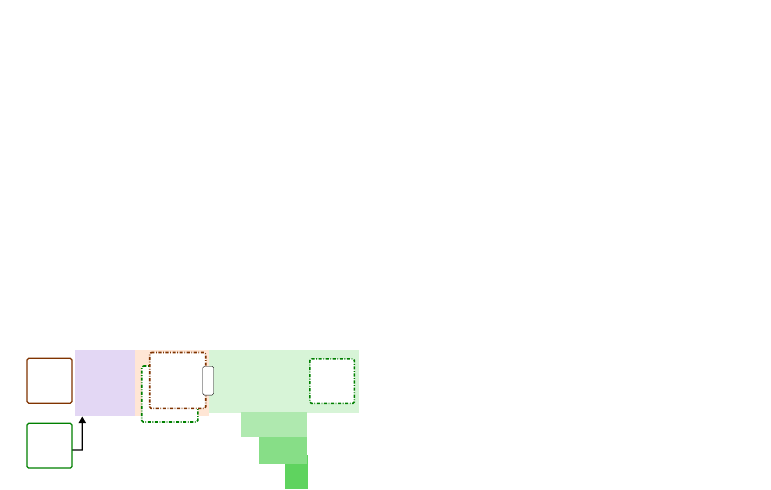
        \vspace{-1.5em}
        \caption{
        The multi-task architecture (color-coded) with a \colorbox[HTML]{e3d7f4}{\color[HTML]{161650}\textit{shared} encoder} (a), its \textit{sequential} variant (b), and the \textit{hybrid} variant (c) for 2 tasks, 
        \colorbox[HTML]{ffe6d5}{\color[HTML]{803300}super-resolution} and
        \colorbox[HTML]{d7f4d7}{\color[HTML]{008000}semantic segmentation}.
        The encoder consists of $B=23$ residual-in-residual blocks~\cite{esrgan2018}, the super-resolution decoder comprises 2 upsampling followed by 2 convolution layers, and the segmentation decoder is taken from HRNet-W18 architecture~\cite{Wang2019hrnet}
        }
        \label{fig:MTPS-scheme}
        \vspace{-1em}
    \end{figure*}

\vspace{-.5em}
\section{Partial multi-task learning}
\label{sec:method}

\subsection{Problem formulation}

Let $\left(L_R, H_R\right)$ be a pair of low- and high-resolution images and
$\left(L_S, M_S\right)$ be a pair of a \textit{different} low-resolution image
with its corresponding semantic annotation. The subscripts $R$ and $S$ indicate
the associated task, respectively super-\underline{r}esolution and semantic
\underline{s}egmentation. Let $c$ be a scale factor, the dual-task 
network is a function $f^c$ producing, for each input image $L_x$ with $N$ pixels,
a semantic mask $\tilde{M}_x$ and a $c$-times higher resolution image $\tilde{H}_x$,
as follows:
\begin{align}
  f^c:\mathbb{R}^N&\rightarrow \mathbb{R}^{c^2N}\times\mathbb{R}^N \\
  L_x&\mapsto \left(\tilde{H}_x,\tilde{M}_x\right).
\end{align}

\vspace{-2em}

\subsection{Network architecture}

The network overview in Figure~\ref{fig:MTPS-scheme}(a)
visualizes with color-code a \colorbox[HTML]{e3d7f4}{\color[HTML]{161650}shared encoder} $E(\cdot)$ for feature extraction and two task-decoder heads, $D^c_R(\cdot)$ and $D_S(\cdot)$ for \colorbox[HTML]{ffe6d5}{\color[HTML]{803300}super-resolution} and \colorbox[HTML]{d7f4d7}{\color[HTML]{008000}semantic segmentation}, respectively.
The predictions are,
 \begin{equation}
    \tilde{H}_x = D_R^c\left(E(L_x)\right) \quad \text{ and} \qquad  \tilde{M}_x = D_S\left(E(L_x)\right).
 \end{equation}
 
\vspace{-.5em}
For super-resolution, we employ the RRDBNet architecture utilized in the ESRGAN~\cite{esrgan2018} model with 23 residual-in-residual dense blocks (RRDB) %
to retain high-level information and increase the encoder's capacity
before upsampling 4 times %
(via interpolations and convolutions) and producing the super-resolved image.
For semantic segmentation, the HRNet~\cite{Wang2019hrnet} architecture is employed, which maintains a stream of embeddings at the input size.
As the RRDBNet encodes the input image while maintaining the resolution of the embeddings, it replaces the first stage of HRNet and acts as the shared encoder $E$, leaving the upsampling subnet the super-resolution decoder $D^c_R$, with 2 pairs of interpolation-convolution layers (for $c=4$) followed by 2 convolution layers. The rest of HRNet becomes the semantic segmentation head $D_S$. Interested readers are referred to the original works for more details. As we do not make any assumption on the architecture, this model is architecture-agnostic.

\paragraph{Multi-task decoders} Different from the \textit{shared}-encoder approach in conventional multi-task models,~\cite{Kim2024SR4IR,Aakerberg2022} proposed a \textit{sequential} approach which concatenates a full HRNet to the RRDBNet such that the super-resolved images are fed directly to the HRNet and uses the segmentation loss to guide the super-resolution performance (Figure~\ref{fig:MTPS-scheme}-(b)). This approach allows maintaining the gradient flow through the super-resolution head when training with segmentation-annotated inputs $L_S$, thereby putting the segmentation performance on the entire dependency of super-resolution. To make the best of both worlds, we propose a \textit{hybrid} head (Figure~\ref{fig:MTPS-scheme}-(c)) where the shared features are adapted and added to the super-resolved features. The three variants share the same encoder $E(.)$ and two decoders $D_R(.)$ and $D_S(.)$, as indicated by the same color code. The added blocks to adapt the features are shown explicitly for clarity purposes. 

\vspace{-.5em}
\subsection{Optimizing objectives} %

The loss function for super-resolution and semantic segmentation task, respectively, can be written as $\mathcal{L}_\text{SR}\left(f(L_R)_{\tilde{H}}, H_R\right)$ and $\mathcal{L}_\text{SS}\left(f(L_S)_{\tilde{M}}, M_S\right)$, which receive different input variables as partial multi-task learning assumes a single-task annotation per training image.
The optimization and parameters update can be performed collectively by alternating image batches for each task as inputs, making two batches per iteration, or separately for each task every epoch. %

\paragraph{Super-resolution task}%
Two losses are employed, including the reconstruction or pixel loss $\mathcal{L}_\text{pix}$ that measures the similarity to the high-resolution target using the $L1$ distance and the Learned Perceptual Image Patch Similarity (LPIPS)~\cite{lpips}
loss $\mathcal{L}_\text{pcp}$ with AlexNet.
We deliberately withhold the adversarial loss, as the ablation study in Table~\ref{tab:ablation-study} shows that it destabilizes the training process and produces suboptimal performance.
The final loss is as follows:

\begin{equation}
\label{eq:loss_SR}
    \mathcal{L}_\text{SR} = \lambda_1\mathcal{L}_\text{pix}(\tilde{H}, H) +
                            \lambda_2\mathcal{L}_\text{pcp}(\tilde{H}, H),
\end{equation}

\paragraph{Semantic segmentation task}%
For the semantic segmentation head, the cross-entropy with softmax loss $\mathcal{L}_\text{CE}$ is employed.
To further constrain the super-resolution head, the super-resolved image is re-projected to the low-resolution domain using nearest neighbor, and a re-projection loss 
is applied with the input image to constrain the fidelity of the output,~\ie~$\mathcal{L}_\text{rpj}(\tilde{H}_S\downarrow_c, L_S)=L1(\tilde{H}_S\downarrow_c, L_S) $.
The total loss for data annotated with semantic segmentation is, therefore:

\vspace{-1em}
\begin{equation}
\label{eq:loss_SS}
    \mathcal{L}_\text{SS} = \mathcal{L}_\text{CE}(\tilde{M}_S,M_S) + \lambda_3\mathcal{L}_\text{rpj}(\tilde{H}_S\downarrow_c, L_S).
\end{equation}

\noindent $\lambda_1$, $\lambda_2$, and $\lambda_3$ are empirically set to 0.01, 1, 4, respectively.

\section{Experiments}
\label{sec:experiments}

\vspace{-1.5em}
\subsection{Setups}

\begin{table} %
    \newcommand{\BlnAak}{SS-guided SR~\cite{Aakerberg2022}}
    \newcommand{\BlnSRIR}{SR4IR~\cite{Kim2024SR4IR}}
    \centering%
    \begin{tabular}{@{}llcc@{}}
        \toprule
                     & variant  &     SSIM &     PSNR  \\                 
        \midrule    
             \BlnAak & sequential &    0.746 &    23.890 \\ %
            \BlnSRIR & sequential &    0.694 &    22.830 \\ %
                Ours & shared     & \bf0.804 & \bf25.491 \\ %
                Ours & hybrid     & \bf0.804 &\ul{25.446}\\ %
        \bottomrule
    \end{tabular}
    \vspace{-.5em}
    \caption{Comparing our shared and hybrid MTL variants with the state-of-the-art methods employing the sequential variant for super-resolution on the Vaihingen test set. Our methods achieve better fidelity quality. 
    \vspace{-2em}
    }
    \label{tab:PMTL-SR}
\end{table}

\begin{table}[t] %
    \newcommand{\BlnAak}{SS-guided SR~\cite{Aakerberg2022}}
    \newcommand{\BlnSRIR}{SR4IR~\cite{Kim2024SR4IR}}
    \centering%
    \begin{tabular}{@{}llcc@{}}
        \toprule
                     & variant  &    Vaihingen & Pleiades \\  %
        \midrule
             \BlnAak & sequential &        0.597 &      0.492 \\ %
            \BlnSRIR & sequential &    \ul{0.716}&  \ul{0.520}\\ %
               Ours  & shared     &        0.605 &      0.508 \\ %
               Ours  & hybrid     &     \bf0.721 &  \bf 0.531 \\ %
        \bottomrule
    \end{tabular}
    \vspace{-0.5em}
    \caption{Comparing our shared and hybrid MTL variants with the state-of-the-art methods employing the sequential variant for semantic segmentation on both datasets. Our methods perform on par with the others on segmentation. 
    \vspace{-1.5em}
    }
    \label{tab:PMTL-SS}
\end{table}

\begin{figure*}[t]
    \centering
    \tiny
    \includeinkscape[width=.285\linewidth]{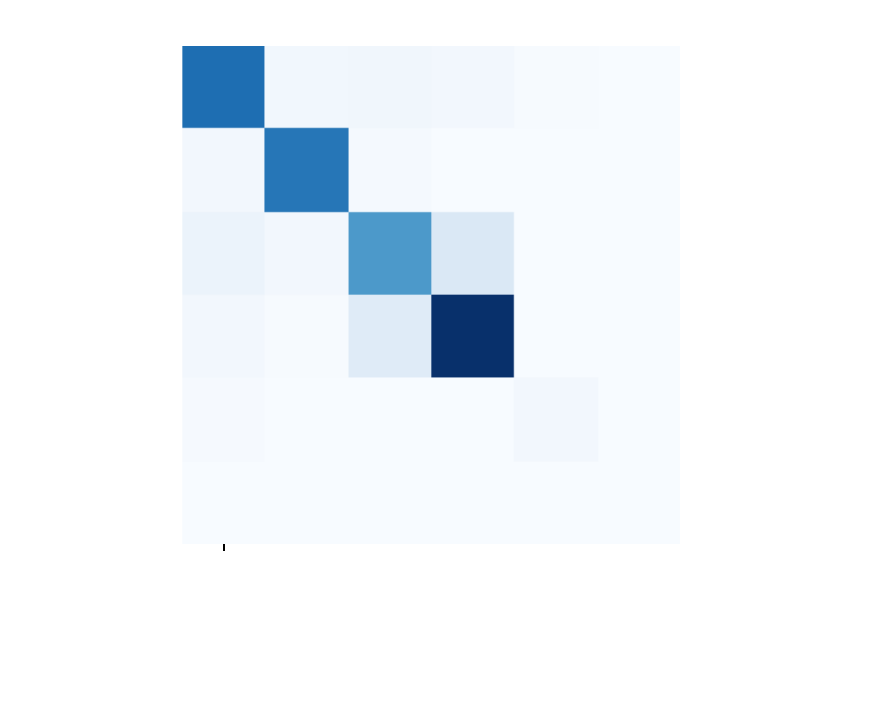}
    \includeinkscape[width=.230\linewidth]{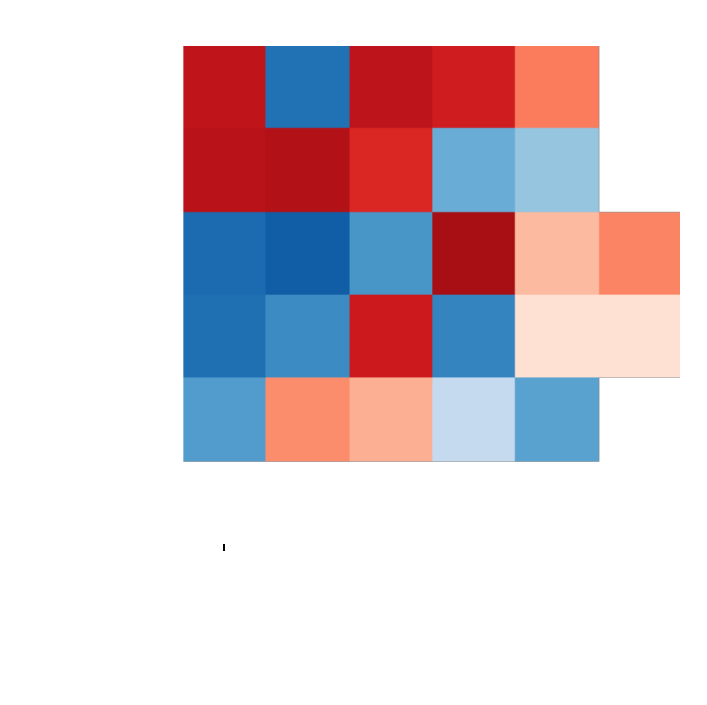}
    \includeinkscape[width=.230\linewidth]{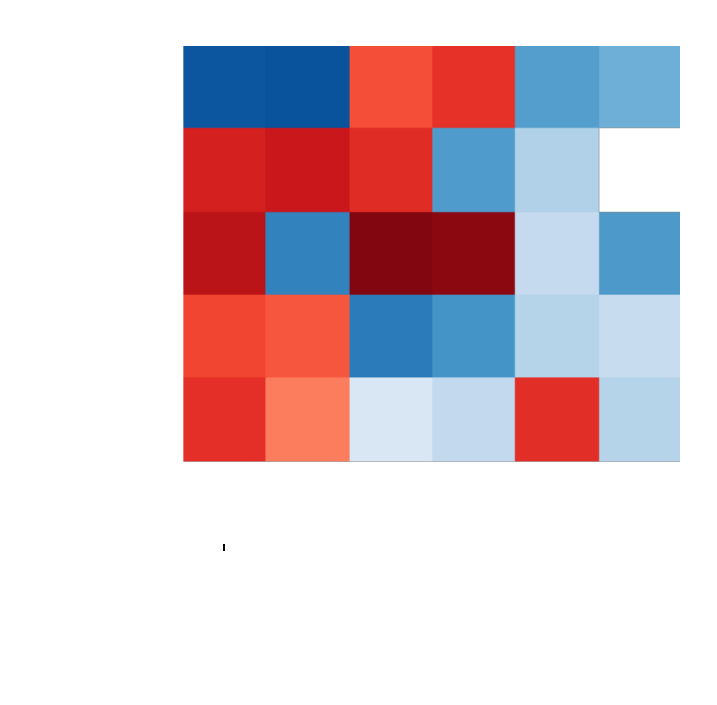}
    \includeinkscape[width=.230\linewidth]{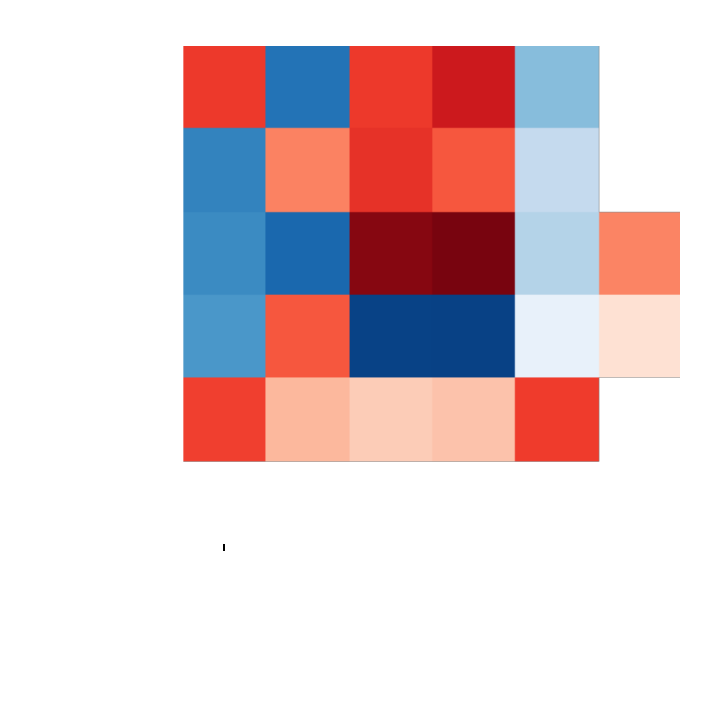}
    \caption{The leftmost figure shows the confusion matrix of SS-guided SR~\cite{Aakerberg2022} ($A$) for the ISPRS data, and the rest show the differences with the confusion matrix ($B$) of each of the other methods: from left to right, SR4IR~\cite{Kim2024SR4IR}, our shared and hybrid variants. The diagonals of the different matrices show if the other methods are higher in true positive counts, calculated by $B - A$, while the other entries indicate if they are lower in false positive counts,~\ie~$A - B$. Positive (red) values are favorable. Both SR4IR and our hybrid approach can recognize the absence of the \texttt{Background} class in the test data (last column).
    }
    \label{fig:confusion_matrix_ISPRS}
\end{figure*}

\begin{figure*}[t]
    \centering
    \newcommand{\BlnAak}{SS-guided SR~\cite{Aakerberg2022}}
    \newcommand{\BlnSRIR}{SR4IR~\cite{Kim2024SR4IR}}
    \begin{tikzpicture}
        \node at (0, 0) {
            \includegraphics[width=\linewidth]{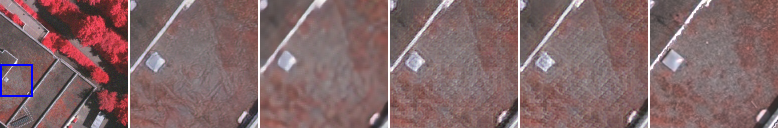}
        };
        \node at (0, 3.00) {
            \includegraphics[width=\linewidth]{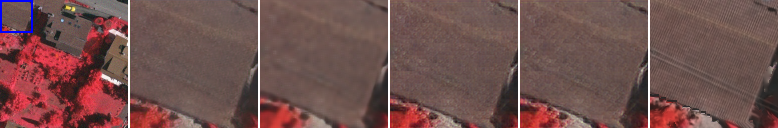}
        };
        \node at (-7.5,4.65) {Input (LR)};  %
        \node at (-4.5,4.65) {\BlnAak}; %
        \node at (-1.5,4.65) {\BlnSRIR};         %
        \node at ( 1.5,4.65) {Ours: shared};         %
        \node at ( 4.5,4.65) {Ours: hybrid};      %
        \node at ( 7.5,4.65) {Target (4x)};       %
    \end{tikzpicture}
    \vspace{-1.5em}
    \caption{Qualitative results on the Vaihingen test. The shared and hybrid variants show richer in textures compared to the overly-smooth SS-guided SR~\cite{Aakerberg2022} and the blurry SR4IR~\cite{Kim2024SR4IR}.
    \vspace{-1em}
    }
    \label{fig:qualitative_isprs}
\end{figure*}

\begin{figure*}[t]
    \centering
    \newcommand{\BlnAak}{SS-guided SR~\cite{Aakerberg2022}}
    \newcommand{\BlnSRIR}{SR4IR~\cite{Kim2024SR4IR}}
    \begin{tikzpicture}
        \node at (0, 0.0) {
            \includegraphics[width=\linewidth]{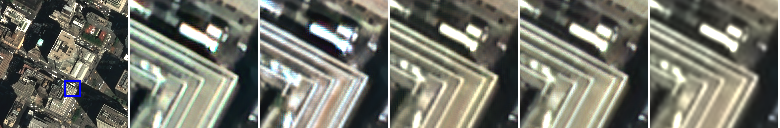}
        };
        \node at (0, 3.00) {
            \includegraphics[width=\linewidth]{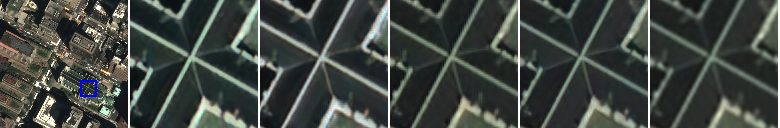}
        };
        \node at (0, 6.00) {
            \includegraphics[width=\linewidth]{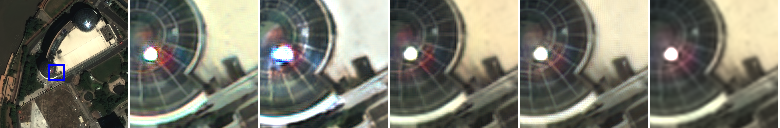}
        };
        \node at (-7.5, 7.65) {Input (LR)};
        \node at (-4.5, 7.65) {\BlnAak};
        \node at (-1.5, 7.65) {\BlnSRIR};
        \node at ( 1.5, 7.65) {Ours: shared};
        \node at ( 4.5, 7.65) {Ours: hybrid};
        \node at ( 7.5, 7.65) {Bicubic};
    \end{tikzpicture}
    \vspace{-1em}
    \caption{Qualitative results on the Pléiades test set. Our methods approach the photometric consistency with the referenced bicubic, but are sharper, while the SOTAs show color leaks and are slightly blurrier. SR4IR~\cite{Kim2024SR4IR} has a strong white-balance bias.
    \vspace{-1em}
    }
    \label{fig:qualitative_cnes}
\end{figure*}

\paragraph{Dataset}
The ISPRS 2D semantic labeling contest~\cite{isprs2012} and the proprietary datasets using real and synthetic Pléiades imagery are employed for experimentation.
Regarding ISPRS, the Potsdam subset is used to train super-resolution and Vaihingen for semantic segmentation and testing.
Following the clean super-resolution (CSR) evaluation strategy~\cite{Lugmayr2019}, the LR images are simulated by degrading clean HR images with hidden kernels and injecting noise of unknown statistics. Inspired by~\cite{Aakerberg2022,BellKligler19,Ji20} we generate $5\times5$-Gaussian kernels $k$ with uniformly random lengths,~\ie~$\sigma_1,\sigma_2\sim\mathcal{U}\left(0.6, 2.5\right)$, and multiplicative noise up to 25\% of the pixels' values,~\ie~$k := k * (1 + 25\%\epsilon), \text{ where }\epsilon\sim\mathcal{N}(0,1)$, to deviate from a regular Gaussian, before normalizing to the sum of 1. The simulated LR images are generated via the well-known model,
\vspace{-.5em}
\begin{equation}
\label{eq:degradation}
   L = \left(H * k\right)\downarrow_c + n,
\end{equation}
\noindent where $n\sim\mathcal{N}(0, 4)$ models the sensor noise. The kernel and noise distribution are kept away from the training process. %
We apply Eq.~\ref{eq:degradation} with the scale ratio $c=4$ to the Vaihingen images whose original ground sampling distance (GSD) is 9~cm, resulting in (LR, HR) pairs of approximately 36~cm and 9~cm GSD. Conversely, the Potsdam images captured at a GSD of 5~cm are first downsampled 2 times using the bicubic kernel to remove potential noise, before applying Eq.~\ref{eq:degradation}, resulting in training pairs of 40~cm and 10~cm GSD. Validating semantic masks from Vaihingen are downsampled by nearest neighbor following common practices.
The Pléiades (CNES/Airbus) dataset\footnote{\scriptsize [Online, accessed 2/2026] Pléiades, VHR (50~cm) Satellites Imagery \url{https://intelligence.airbus.com/imagery/our-optical-and-radar-satellite-imagery/pleiades/}.} is set as the target data with the scale factor of 2, including 230 pan-sharpened images and semantic masks of $2048\times2048\times4$-resolution and 50~cm GSDs (R, G, B, and Near-IR). For training super-resolution, a synthetic dataset with representative characteristics of the Pléiades sensor, such as spectral bands, FTM, and SNR, is used, comprising 95 LR-HR pairs with the GSDs of 50~cm and 25~cm, respectively.

\paragraph{Metrics}
The Peak Signal-to-Noise Ratio (PSNR) and Structure Similarity Index (SSIM) are used for super-resolution, and mean intersection-over-union (mIoU) for semantic segmentation, of which higher is better.
As Pléiades is collected in a real-world setting with no paired HR images, the SR results are presented qualitatively only.

\subsection{Partial multi-task learning}

In this subsection, the proposed variants for partial multi-task learning (PMTL) are compared to the state-of-the-art approaches using the sequential variant, namely SR4IR~\cite{Kim2024SR4IR} and SS-guided SR~\cite{Aakerberg2022}. To maintain comparability, we follow each paper's description and re-implement using our codebase. In particular, we respect the choices for losses and their empirical weights, as well as initialize all networks from pretrained models. For SR4IR~\cite{Kim2024SR4IR}, the CQMix augmentation is removed for semantic segmentation as it violates the assumption of multi-task partially supervised learning. All networks are trained for 200K iterations with early stopping at the highest mIOU.

\paragraph{For super-resolution} the quantitative results on ISPRS test set are shown in Table~\ref{tab:PMTL-SR}. The PMTL approaches outperform both state-of-the-art methods on structure similarity (SSIM) and achieve lower distortion (PSNR), showing higher image quality. The qualitative results on ISPRS in Figure~\ref{fig:qualitative_isprs} show better textures in our methods compared to the overly smooth SS-guided SR~\cite{Aakerberg2022} and blurry SR4IR~\cite{Kim2024SR4IR}.
On Pléiades images (Figure~\ref{fig:qualitative_cnes}) and with reference to the bicubic-upsampled version, our results are closer in photometric consistency, especially at brighter areas, where other results contain color leaks in brighter areas due to strong white balance bias that overly whitens the details.

\paragraph{For semantic segmentation} in Table~\ref{tab:PMTL-SS}, the hybrid and SOTA-method SR4IR~\cite{Kim2024SR4IR} experience a jump in performance on Vaihingen compared to the other two. Inspection of the confusion matrices in Figure~\ref{fig:confusion_matrix_ISPRS} shows that the \texttt{Background} class is absent from the Vaihingen test set, which is recognized by both approaches (in such a case, the common practice disregards the class from mIoU calculation), while the others produce false positive results, which dictates the class being used in the calculation. Both our variants produce similar performance to the SOTA approaches with slight numerical benefits on both Vaihingen and Pléiades datasets.

\subsection{Ablation study}
\label{sec:ablation}

Table~\ref{tab:ablation-study} provides an ablation study of the GAN and re-projection $\mathcal{L}_\text{rpj}$ losses for all 3 variants. All three networks are trained from scratch and employ the same set of losses. Thus, the performance does not match the SOTA comparison in the previous section.
Although $\mathcal{L}_\text{rpj}$ does not provide extra information, it guarantees the fidelity of super-resolved images and reduces distortions. The GAN loss, on the other hand, destabilizes the training process and is disadvantageous to image quality.

\begin{table}[t]
    \newcommand{\BlnG}{sequential}%
    \newcommand{\wGAN}{\small with GAN}
    \newcommand{\wLprj}{\small with $\mathcal{L}_\text{rpj}$}
    \setlength{\tabcolsep}{1.7pt}
    \centering%
    \begin{tabular}{@{}cccc@{}}
        \toprule
                  &                             &   \multicolumn{2}{c}{sequential}  \\
        \cmidrule(lr){3-4}  
        \wGAN & \wLprj                          &     SSIM &      PSNR  \\ %
        \midrule    
        \checkmark&\checkmark                   &    0.796 &     26.000 \\ %
                  &                             &    0.764 &     23.835 \\ %
                  &\checkmark                   & \bf0.821 & \bf 26.491 \\ %
        \bottomrule
    \end{tabular}
    \begin{tabular}{@{}cc@{}}
        \toprule
                                                    \multicolumn{2}{c}{shared}  \\
        \cmidrule(lr){1-2}  
                                                      SSIM &      PSNR  \\ %
        \midrule    
                                                     0.778 &   25.434 \\ %
                                                     0.787 &   24.657 \\ %
                                                \bf 0.822 &\bf26.413 \\ %
        \bottomrule
    \end{tabular}
    \begin{tabular}{@{}cc@{}}
        \toprule
        \multicolumn{2}{c}{hybrid-add}                            \\
        \cmidrule(lr){1-2}
                                          SSIM     &      PSNR  \\
        \midrule                           
                                             0.767 &   24.971 \\ %
                                             0.767 &   24.073 \\ %
                                        \bf  0.827 &\bf26.634 \\ %
        \bottomrule
    \end{tabular}
    \caption{Ablation study of GAN loss and re-projection loss $\mathcal{L}_\text{rpj}$. Removing GAN loss while adding re-projection loss reduces super-resolution distortion.}
    \vspace{-1em}
    \label{tab:ablation-study}
\end{table}

\vspace{-1em}
\section{Conclusion}
\label{sec:conclusion}

The paper studies the joint learning under the multi-task partially supervised framework for Earth observation imagery.
It is based on the well-known assumption that the two tasks are highly relevant and could contribute to one another. The hybrid architecture is proposed to inherit the best of both shared representation and the semantic-guided sequential variants. The experiments against SOTA sequential methods show higher super-resolution quality both quantitatively and visually, as well as similar semantic segmentation performance with a slight numerical benefit.

\clearpage

\bibliographystyle{IEEEbib}
\bibliography{macro,IRISA-ACCV24}

\end{document}